%% file: example_paper.tex
\documentclass{article}

\usepackage{microtype}
\usepackage{graphicx}
\usepackage{tabularx}
\usepackage{subcaption}
\usepackage{booktabs} 
\usepackage{multirow}
\usepackage{array}

\usepackage{hyperref}

\usepackage[preprint]{icml2026}

\usepackage{amsmath}
\usepackage{amssymb}
\usepackage{mathtools}
\usepackage{amsthm}

\usepackage{xcolor}
\usepackage{ifthen}

\usepackage[capitalize,noabbrev]{cleveref}

\theoremstyle{plain}

\theoremstyle{definition}

\theoremstyle{remark}

\newboolean{showcomments}
\setboolean{showcomments}{true}  

\newcommand{\authorcomment}[3]{%
  \ifthenelse{\boolean{showcomments}}%
  {{\color{#2}\textbf{[#1:} #3\textbf{]}}}%
  {}%
}

\icmltitlerunning{Em-Garde: A Propose-Match Framework for Proactive Streaming Video Understanding}

\begin{document}

\twocolumn[
  \icmltitle{Em-Garde: A Propose-Match Framework for Proactive Streaming Video Understanding}



  \icmlsetsymbol{equal}{*}

  \begin{icmlauthorlist}
    \icmlauthor{Yikai Zheng}{air,iiis,equal}
    \icmlauthor{Xin Ding}{ustc,equal}
    \icmlauthor{Yifan Yang}{ms}
    \icmlauthor{Shiqi Jiang}{ms}
    \icmlauthor{Hao Wu}{nju}
    \icmlauthor{Qianxi Zhang}{ms}
    \icmlauthor{Weijun Wang}{air}
    \icmlauthor{Ting Cao}{air}
    \icmlauthor{Yunxin Liu}{air}
  \end{icmlauthorlist}

  \icmlaffiliation{air}{Institute for AI Industry Research (AIR), Tsinghua University}
  \icmlaffiliation{iiis}{Institute for Interdisciplinary Information Sciences (IIIS), Tsinghua University}
  \icmlaffiliation{nju}{Nanjing University}
  \icmlaffiliation{ustc}{University of Science and Technology of China}
  \icmlaffiliation{ms}{Microsoft Research Asia}
  \icmlcorrespondingauthor{Ting Cao}{tingcao@mail.tsinghua.edu.cn}

  \icmlkeywords{Machine Learning, ICML}
  \vskip 0.3in
]



\printAffiliationsAndNotice{}  

\begin{abstract}
  Recent advances in Streaming Video Understanding has enabled a new interaction paradigm where models respond proactively to user queries. Current proactive VideoLLMs rely on per-frame triggering decision making, which suffers from an efficiency-accuracy dilemma. We propose \textbf{Em-Garde}, a novel framework that decouples semantic understanding from streaming perception. At query time, the Instruction-Guided Proposal Parser transforms user queries into structured, perceptually grounded visual proposals; during streaming, a Lightweight Proposal Matching Module performs efficient embedding-based matching to trigger responses. Experiments on StreamingBench and OVO-Bench demonstrate consistent improvements over prior models in proactive response accuracy and efficiency, validating an effective solution for proactive video understanding under strict computational constraints. Code is available at \url{https://github.com/air-embodied-brain/Em-Garde}
\end{abstract}













\section{Introduction}
\begin{figure*}[t]
    \centering
    \includegraphics[width=\linewidth]{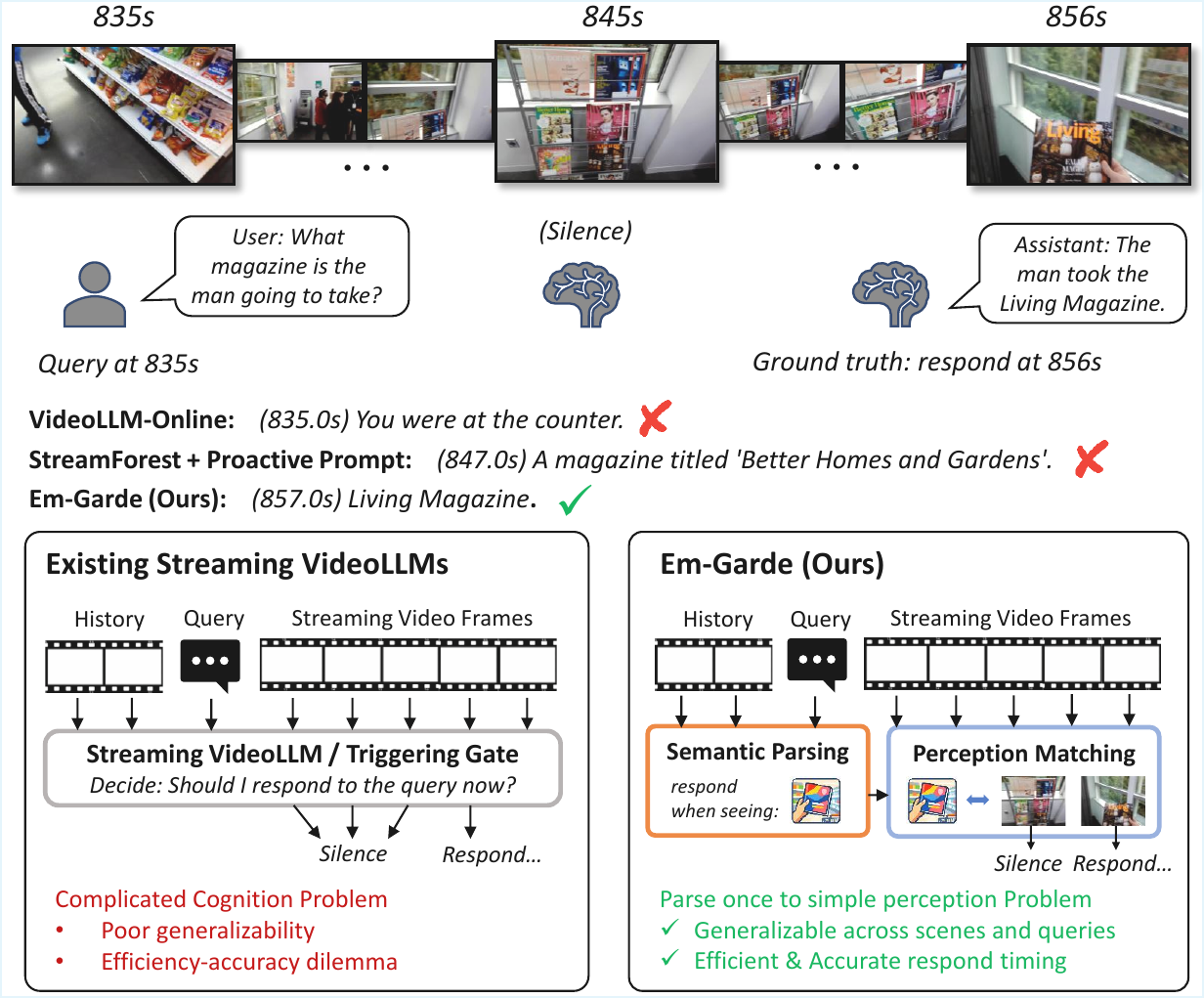}
    \caption{Demonstration of our model v.s. existing Streaming VideoLLMs on the Proactive Streaming Understanding task. While existing models solve a complicated response/silence decision-making problem at every timestep, we turn the problem into a simple perception problem with query-time semantic parsing, allowing for efficient and accurate proactive response timing.}
    \label{fig:teaser}
\end{figure*}
The recent advances in Multimodal LLMs (MLLMs)~\cite{qwen2_5, videollama, llava-onevision, llavavideo, internvl2, longvu, yao2024minicpm, liu2024llavanext}, have brought about significant success in streaming video understanding. Industrial models like Gemini Live~\cite{team2024gemini} are able to comprehend streaming video content and respond to user queries in real time, bringing an unprecedented interactive experience. Academic efforts push even further towards models that can answer a user query proactively~\cite{vl-online, vl-mod, streammind, streambridge, dispider, livecc}: instead of passively switching on and answering at query time, these models keep monitoring the video stream, and automatically decide when to respond to the query (e.g. remind the user when water boils). This new interactive paradigm opens up vast new applications from sports commentary to household assistance~\cite{SoccerNet, livecc, xu2025streamingvlm, Merrill2022AICF, wang2025streameqa, wen2025aiservice, li2025redundancy}, yet also posing a fundamental new challenge to the community.


The unique challenge of proactive streaming video understanding lies in the need to decide whether to respond in real time at every timestep~\cite{mmduet, vl-online, dispider}. Whenever a new frame arrives, the system must perform multiple perceptual and semantic reasoning processes to determine whether a response should be triggered, including visual perception (e.g. object and action recognition), event transition detection, and query-aware relevance assessment. However, the processing speed must match the video frame rate (at least 5-10 frames per second (fps)), making the available computational budget severely limited~\cite{rekv, timechat, qian2024streaming,edgemind}. This creates an inherent tension between the need for \emph{complicated visual-semantic understanding and the requirement for fast per-frame computation}.

Existing approaches mostly formulate the problem as a \textbf{per-frame decision making problem}. Specifically, they utilize an MLLM or lightweight gate to output a response/silence decision at every timestep. To keep up with the streaming speed, they reduce the model size~\cite{streammind, streambridge, dispider} or increase information compression rate~\cite{fvstream, qian2024streaming, zeng2025streamforest, timechat, wang2025stc, ning2025livevlm} to perform fast framewise understanding. However, this naturally limits the granularity of visual perception and the decision-making quality of the model, making the tension between the need for rich visual understanding and the strict real-time constraint fundamentally unsolved.


To solve the efficiency-accuracy dilemma, we try to break away from this paradigm. We note that the complicated decision-making problem can break down into two stages: Upon the query, the model can already decide which kind of events are worth responding to and imagine the visual evidence that points to them. Then, only simple visual perception is needed to match the evidence~\cite{lee2025flashback, yang2025streamagent}. Based on this insight, we propose a \textbf{per-query parsing paradigm} (\cref{fig:teaser}) that shifts expensive semantic interpretation entirely outside the streaming loop. Specifically, the system parses the user query into perceptually grounded proposals at query time, enabling subsequent streaming processing to be reduced to efficient frame-level visual matching. For example, if the user requests to be notified when “the water boils”, the system does not need to repeatedly interpret this request at every frame. Instead, it parses the instruction at query time into concrete visual proposals, such as “vigorous bubbling” or “sustained steam emission”, and then continuously performs lightweight frame-level matching to detect these cues.

This separation introduces two key technical challenges. Firstly, there is no universal parsing scheme that can generalize across diverse and open-ended user queries. The system must handle varied query structures, multiple possible event types, and different levels of abstraction, and produce structured visual proposals that a lightweight perception model can reliably use. Secondly, given multiple fine-grained visual proposals, it is non-trivial for a lightweight model to quickly and reliably match them to actual visual evidence in streaming video.

By addressing these challenges, we realize a novel framework for proactive streaming video understanding, called \textbf{Em-Garde}, which separates semantic reasoning from perception and enables real-time proactive response under strict computational constraints. It integrates two key techniques. 

For semantic parsing, we introduce the Instruction-Guided Proposal Parser (IGPP), which leverages the reasoning ability of the MLLM to parse high-level instructions into structured visual proposals. Each proposal consists of parallel visual cues describing either distinct noteworthy events or detailed aspects of a single event. IGPP incorporates a short video context to adapt proposals over time, enabling the parsing process to evolve with the ongoing video while still respecting the query instruction. 

To train IGPP, we curate the Parse2Prop-1K dataset containing pairs of queries, proposals, and target response times written by humans or GPT-5~\cite{singh2025openaigpt5card}, covering diverse proposing methods, variable proposal lengths, and different levels of abstraction.

For streaming perception, we propose the Lightweight Proposal Matching Module (LPMM). LPMM encodes recent video context with a sliding window and embeds both the video segment and the IGPP proposals using a lightweight embedding model~\cite{zhai2023sigmoid, meng2025vlm2vec, zhang2024gme, jian2025rzenembed}. The embeddings are compared in the embedding space to produce a temporal sequence of similarity scores. A surge in similarity indicates a potential matching, triggering the MLLM responder to generate a response and verify its appropriateness. This design decouples expensive reasoning from high-frequency perception, enabling efficient, frame-rate-matched proactive responses. 

To our knowledge, Em-Garde is the first proactive streaming framework that explicitly decomposes triggering decision into query-time proposal generation and streaming-time proposal matching, thus reducing the complex streaming understanding problem to lightweight perception problem.

To evaluate the performance of the Em-Garde framework, we perform extensive tests on proactive response tasks and online video understanding tasks. On proactive response tasks, we outperform existing real-time streaming models by more than 3\% accuracy on StreamingBench~\cite{lin2024streamingbench} and 10\% F1 score on OVO-Bench~\cite{ovo}, while achieving state-of-the-art processing speed of 10-15 fps on A100 GPUs on arbitrarily long videos~(\cref{fig:efficiency}). On online video understanding tasks, we score an overall 76.7\% on StreamingBench real-time perception tasks, 63.0\% on OVO-Bench real-time perception tasks, and 52.2\% on OVO-Bench backward tracing tasks. Further analysis shows that the quality of the IGPP proposal contributes to the trigger accuracy and that RL helps generate perception-friendly proposals. These results show the efficiency and generalizability of our framework on the proactive streaming understanding task.

\section{Related Work}
The concept of an always-on, proactive streaming assistant was first articulated in VideoLLM-Online~\cite{vl-online}, which introduces a conversational setting where users provide a query in advance and the model autonomously responds when relevant events occur. This interaction paradigm has since been widely adopted in subsequent works~\cite{lionfs, streambridge, streammind, dispider, xia2025streamo, fu2025vispeak, zhang2025eyeswideopenego, zhang2025avila}, establishing proactive response as a core problem in streaming video understanding.

A central challenge in proactive streaming is determining when to respond, often referred to as the \textbf{triggering decision} problem. VideoLLM-Online~\cite{vl-online} trains the model to output a silence or respond token after each frame, continuing generation only if the respond token is produced. StreamMind~\cite{streammind} and Dispider~\cite{dispider} follow this paradigm but introduce a separate gating module to improve efficiency. However, due to the difficulty of collecting high-quality supervision for triggering signals, such models often suffer from limited generalizability and struggle to adapt to diverse or unseen queries.

An alternative strategy, introduced in OVO-Bench~\cite{ovo}, prompts a VideoLLM to continuously predict whether a target event has occurred, triggering a response when the prediction changes from “No” to “Yes.” Streaming VideoLLMs such as FlashVStream~\cite{fvstream} and StreamForest~\cite{zeng2025streamforest} adopt this scheme and focus on improving efficiency through context management. Nevertheless, VideoLLMs are known to follow such proactive prompts unreliably~\cite{lin2024streamingbench}. Moreover, the triggering decision frequently requires fine-grained visual–semantic reasoning, which conflicts with the aggressive information compression typically employed by streaming models to maintain real-time performance.

More recent approaches, including StreamAgent~\cite{yang2025streamagent} and MMDuet-2~\cite{wang2025mmduet2}, recognize that determining the appropriate response time requires non-trivial reasoning over evolving visual evidence. Instead of solving it with naive supervised training or prompting, they formulate proactive response as a sequential decision-making problem and address it via reinforcement learning or step-wise planning. Although these methods improve accuracy, their reliance on long contexts and frame-wise reasoning introduces substantial computational overhead, making real-time deployment challenging.

Overall, existing methods attempt to make triggering decisions from scratch at every timestep. This leaves unresolved the fundamental tension between the rich visual–semantic understanding required for accurate triggering and the limited computational budget available in streaming settings. In contrast, we move semantic parsing outside the streaming loop, thereby reducing the computational burden of triggering decisions. The streaming loop can then rely on a lightweight perception model operating independently to detect relevant visual signals, without repeated semantic reasoning. By decoupling query understanding from visual perception, our approach enables accurate, real-time proactive responses with substantially lower computational overhead.

\section{Methodology}
\begin{figure*}[t]  
    \centering
    \includegraphics[width=\textwidth]{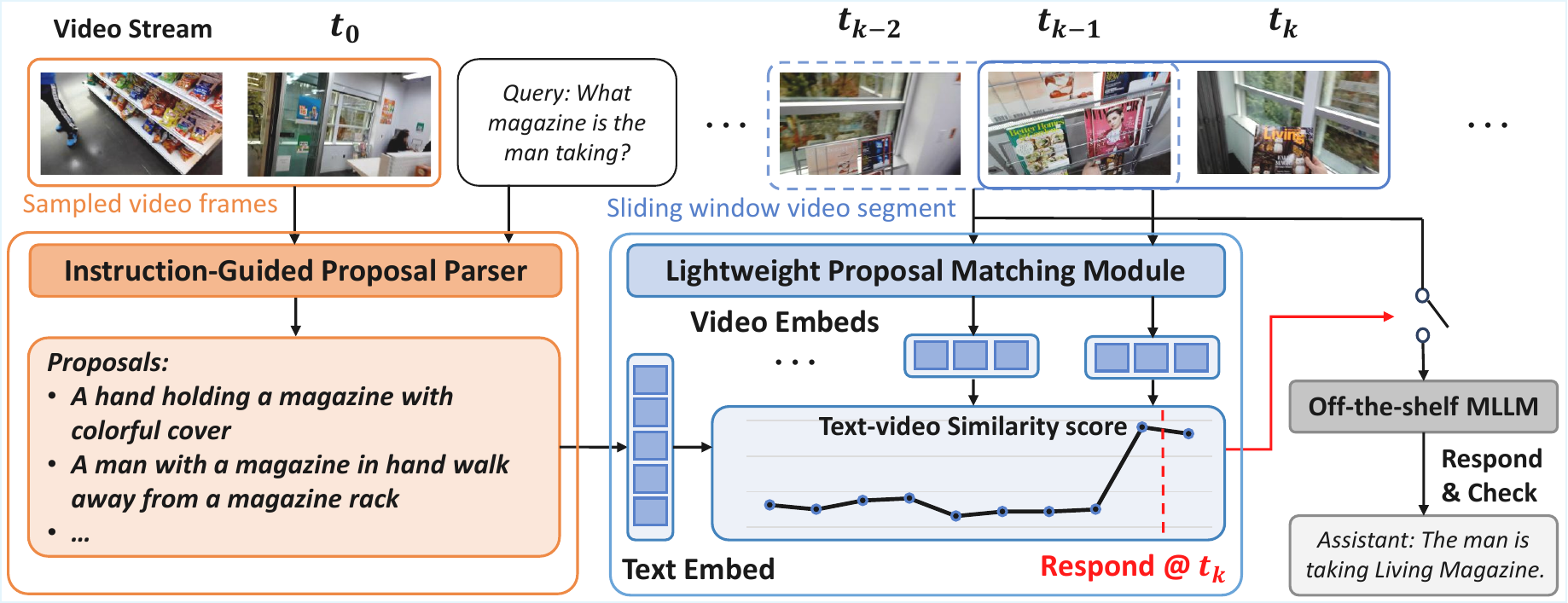}  
    \caption{\textbf{Overview of the Em-Garde Framework:} \textbf{\color{orange}IGPP (Orange)} receives the Instrcution $I$ and a low-fps video context before query time, and parse the instruction into perceptually-grounded visual cues. \textbf{\color{blue}LPMM (Blue)} runs in the streaming loop, matching the current sliding-window video segment to the proposal in the embedding space. The similarity scores are utilized as the temporal signal for response triggering decision. Together, the two modules separate semantic understanding from the streaming loop, allowing for efficient and accurate response time decision.}
    \label{fig:principal}
\end{figure*}
\subsection{Problem Formulation}
We adopt the standard formulation of proactive streaming video understanding introduced in prior work \cite{vl-online, dispider}. In this setting, a model receives a natural language instruction once at query time and subsequently processes a continuous video stream, deciding when to respond without further user queries.

Formally, an instruction $I$ is provided at query time $t_0$. A video stream is represented as an unbounded sequence of frames $\{x_t\}_{t\geq 0}$ where frame $x_t$ arrives at time step $t$. At each time step $t\geq t_0$, the model observes the video frames until $t$ and outputs a binary triggering decision 
\begin{equation}
    D_t=\pi_{trigger}(I, \{x_\tau\}_{0\leq \tau\leq t})\in\{0,1\},
\end{equation}
Where $D_t=1$ indicates that a response should be issued at time $t$.

Suppose the model decides to respond at time step $t_1$, i.e. $D_{t_1}=1$, the model generates a response 
\begin{equation}
    R_{t_1}=\mathcal{G}(I, \{x_{\tau}\}_{0\leq\tau\leq t_1}),
\end{equation}
which may take the form of an answer, alert, explanation, or tool invocation, depending on the instruction. 

Let \begin{equation}
    T_R=\{t|D_t=1\}
\end{equation}
denote the set of predicted response times. Performance is evaluated by matching predicted triggers to ground-truth event times within a tolerance window, following established benchmark protocols (e.g., OVO-Bench, StreamingBench).

\subsection{The Overview of Em-Garde}
To improve the accuracy and efficiency of triggering decisions, we decompose the decision process into two stages: semantic parsing and visual perception. Formally, given an instruction $I$ and video frames $\{x_\tau\}_{t_0-h_c\leq \tau\leq t_0}$ of length $h_c$ around the query time, a proposer model generates a proposal set:
\begin{equation}
    P=\pi_{propose}(I,\{x_\tau\}_{t_0-h_c\leq \tau\leq t_0}).
\end{equation}

The proposal describes observable visual evidence about the events that should trigger a response according to the instruction. This transforms the original triggering decision problem into a perception-centric matching problem. 

After semantic parsing, a lightweight visual matching module runs continuously in the streaming loop. At each time step $t$, the module receives a short streaming video segment $\{x_\tau\}_{t-h_s\leq\tau\leq t}$ of length $h_s$ and the proposal set $P$, and outputs a set of matching score
\begin{equation}
    S^t= \pi_{match}(P,\{x_\tau\}_{t-h_s\leq\tau\leq t}).
\end{equation}
Triggering decisions are then derived via simple temporal processing over the score history:
\begin{equation}
    D_t = \pi_{tp}(\{S^\tau\}_{\tau\leq t}).
\end{equation}

Based on this formulation, we proposed \textbf{Em-Garde} (\cref{fig:principal}), a proactive streaming understanding framework that explicitly separates semantic parsing from visual perception. Em-Garde consists of two key components:
\begin{itemize}
    \item \textbf{Instruction-Guided Proposal Parser (IGPP)}: IGPP is invoked once at query time. It performs the complex reasoning to translate high-level instructions into concrete perceptually-grounded cues, effectively defining a decision schema for triggering responses. 
    \item \textbf{Lightweight Proposal Matching Module (LPMM)}: LPMM runs continuously in the streaming loop. It matches incoming video segments against the proposal set in a multimodal embedding space and produces temporal matching scores without performing full semantic understanding of the video.
\end{itemize}
This design achieves efficient allocation of computational budget and avoids repeated semantic understanding, preserving task fidelity while enabling real-time performance.
\subsection{Instruction-Guided Proposal Parser}
At query time $t_0$, IGPP receives the instruction $I$ along with a segment of video history $\{x_\tau\}_{t_0-h_c\leq \tau\leq t_0}$. Using a large multimodal language model, it generates a proposal set: 
\begin{equation}
    P=\{p_1,p_2,\dots,p_k\}
\end{equation}
Each proposal $p_i$ is a \textbf{concise, declarative} visual cue describing a response-worthy event.
\subsubsection{Properties of the proposals} Proposals are intentionally designed to have the following properties:
    
\begin{itemize}
    \item \textbf{Temporal Localizability} Proposals should only be matched when a response is required. 
    \item \textbf{Perceptual Groundability} Proposals can be matched with a short video segment by simple visual perception without understanding the long context.
    \item \textbf{Redundancy} Multiple proposals are allowed to describe complementary visual details of the same underlying event.
\end{itemize}

Property 1 and 2 follows naturally from the role of IGPP to turn response/silence decision-making into simple perception problem, while Property 3 reflects the inherent uncertainty of a streaming video. Rather than committing to a single hypothesis, IGPP maintains a set of plausible evidences.


\subsubsection{Learning effective proposals} Generating useful proposals is non-trivial, involving instruction understanding and task-specific
reasoning. Naively trained models might produce proposals that the perception module cannot reliably recognize.

To address this, we train IGPP in two stages. First, \textbf{supervised fine-tuning (SFT)} encourage the model to learn the proposal format and pre-defined proposal methods. Then, we introduce a \textbf{reinforcement learning (RL)} stage that directly optimizes downstream triggering behavior. In this stage, the model tests out the learned proposal methods, aligns the visual cues with the perception module, and learns to generate proposal sets that are temporally localized and perceptually grounded.

For training data, we curated \textbf{Parse2Prop-1K}, a small-scale dataset containing 668 queries and proactive responses on 92 videos across COIN~\cite{coin}, Ego4d~\cite{grauman2022ego4d} and BEHAVIOR~\cite{li2024behavior1k}. Each query is issued at time $t_0$ prior to the occurrence of the target event and may require one or multiple responses after $t_0$. Half of the examples include human-or-GPT-5-authored proposal annotations. The proposals exhibit diverse styles, lengths, and levels of visual detail, with an average of 3.99 visual cues per query and an average cue length of 13.5 words. Details of the dataset can be found in~\cref{sec:dataset} of the appendix.

During reinforcement learning, the proposer is rewarded for correct triggering decisions near the onset of each ground-truth event, receiving one point per correct trigger. A trigger is considered correct if it is within a tolerance window after the start of an event (4 seconds in our experiments), accounting for potential information-gathering latency. Triggers outside these windows are treated as false positives and penalized. A coefficient $\lambda$ is introduced to balance between event recall and false trigger suppression. The final reward is defined as

$$r=(1-\lambda r_{fp})\frac{n_c}{n}$$

Where $$r_{fp}=1-2^{-n_{fp}/n}$$

and $n_c$, $n_{fp}$, $n$ are the number of correct triggers, false triggers, and ground-truth events, respectively. We set $\lambda=1$, and train the model using GRPO \cite{shao2024deepseekmath}.


\subsection{Lightweight Proposal Matching Module}
LPMM is instantiated as a VLM-based multimodal embedding model~\cite{zhang2024gme,meng2025vlm2vec}. During streaming, at each time step $t$, a short sliding window of recent frames $\{x_\tau\}_{t-h_s\leq \tau\leq t}$ is embedded into a video representation $e_v(t)$. Unlike IGPP, LPMM is not required to perform semantic reasoning; instead, it focuses on \textit{detecting immediate visual cues} relevant to the proposal set. Therefore, it uses a short temporal context with a higher frame rate to capture time-sensitive actions and object appearances.

Each proposal $p_i \in P$ is pre-embedded into a proposal representation $e_{p_i}$. The matching score between the video segment and proposal $p_i$ at time $t$ is computed using cosine similarity:
\begin{equation}
    s_i^t=\cos(e_v^t,e_{p_i})
\end{equation}

Triggering decisions are derived from the temporal evolution of similarity scores. Specifically, a trigger occurs when at least one proposal exhibits a sharp increase in similarity exceeding a predefined threshold: $$D_t=\mathbf{1}\big{[}\max_i (s_i^t-s_i^{t-1})>\theta\big{]}$$

The threshold $\theta$ governs the \textit{trade-off between triggering sensitivity and conservativeness}. Increasing $\theta$ generally reduces spurious triggers but also increases the risk of missed events, which may be unacceptable in safety-critical applications such as healthcare and surveillance. Conversely, lower $\theta$ values favor more frequent triggering, improving recall at the cost of increased downstream verification, which may be undesirable in daily-life scenarios such as household assistance. Em-Garde exposes $\theta$ as an explicit control knob, allowing practitioners to select operating points appropriate to application requirements.

Since LPMM does a general perception task which does not involve task-specific instruction followings, we employ an off-the-shelf embedding model, Ops-MM-V1~\cite{opsmm}, for the LPMM without any training.

\subsection{Computational Efficiency}
By shifting semantic understanding out of the streaming loop, Em-Garde runs the streaming loop with a lightweight model on constant length input, inherently removing the problem of linearly growing inference time without applying aggressive information compression.

 To further improve the inference efficiency, we introduce a visual encoding cache, which stores the encodings of the overlapped video frames in the sliding window. This allows the visual encoder in the embedding model to encode only one video frame at every timestep, further improving the efficiency by $2\times-3\times$. With these improvements, the streaming loop can run at a maximum of 10-15 fps on arbitrarily long video sessions on A100 GPUs.

The proposal and response generation, while more computationally expensive, are executed infrequently and can be scheduled asynchronously. Consequently, they do not block or degrade the efficiency of the main streaming loop, yielding a scalable and practical solution for long-horizon proactive streaming understanding.
\section{Experiments}
\subsection{Experiment Setup}
\subsubsection{Implementation Details} Our proposal model is initialized from Qwen2.5VL-7B~\cite{qwen2_5} and trained with our two-stage framework on 8 A100 GPUs. The SFT and RL stages took around 1 and 5 hours respectively. We use Ops-MM-V1-2B~\cite{opsmm} as the embedding model. During inference, the embedding model embeds 2-second, 2fps video segments at a process rate of 2fps. The proposal model receives the original query together with 5 seconds of video history at 1 fps as the context. The triggering threshold is set to 0.04 by default.
\subsubsection{Benchmarks for Evaluation}
With our experiments, we aim to evaluate three key aspects in proactive streaming understanding: 
\begin{itemize}
    \item Whether our model yields superior \textbf{proactive response} abilities, i.e. the ability to decide when to respond to a user query.
    \item Whether our model preserves \textbf{online video question-answering} abilities of Streaming VideoLLMs, which evaluate the quality of the responses. 
    \item Whether our model runs streaming inference \textbf{efficiently}. 
\end{itemize}
For Proactive Response, we selected three popular benchmarks: The Proactive Response (PO) task of streamingbench~\cite{lin2024streamingbench}, The Forward Active Response (FAR) task of OVO-Bench~\cite{ovo}, and ProactiveVideoQA~\cite{wang2025proactivevideoqa}. For Online Video QA, we selected the Real-time Understanding task of StreamingBench, and the Real-time Perception and Backward Tracing tasks of OVO-Bench.
\input{tables/ovo_online}
\input{tables/streambench_online}
\input{tables/proactivevideoqa}
\subsection{Results on Proactive Response tasks}
We first evaluate our model’s proactive streaming understanding ability on three benchmarks: StreamingBench, OVO-Bench, and ProactiveVideoQA.

For StreamingBench and ProactiveVideoQA, we follow the original evaluation metrics (accuracy for StreamingBench and PAUC for ProactiveVideoQA). For OVO-Bench, proactive models were scarce when the benchmark was released, and it therefore adopted an indirect offline accuracy metric. However, we observe that this metric is not always aligned with the actual correctness of triggering decisions (see~\cref{sec:ovo_proactive} of the appendix). We therefore return to the benchmark’s original evaluation goal and adopt an \textbf{Online Recall and Precision} metric:

\begin{equation}
Rec=\frac{|T_c|}{|T_{gt}|},\quad
Pre=\frac{|T_c|}{|T_r|}
\end{equation}

Here, $T_{gt}$ denotes the set of ground-truth response times and $T_r$ the set of model response times. $T_c$ represents the set of correct response times, defined as responses that fall within a tolerance window of a ground-truth response time.

For OVO-Bench, we evaluate existing streaming models under this online protocol using each model’s original proactive response mechanism. In particular, for models that require proactive prompting (FVStream and StreamForest), we follow the prompting method provided by OVO-Bench but evaluate triggering correctness using our online metric instead of the original offline accuracy. As Dispider and StreamAgent have not released their proactive inference code, they are not included in our evaluation.

The results are shown in \cref{tab:ovo-online}, \cref{tab:po}, and \cref{tab:proactiveqa}. For each benchmark, we compare with baselines with reported results or available implementation. Our method significantly outperforms all baselines on OVO-Bench and StreamingBench while achieving competitive performance on the newer and less-tested ProactiveVideoQA benchmark, demonstrating the strong proactive response capability of our model.

\subsection{Results on Online Video Understanding tasks}

To evaluate the online video question-answering abilities of our framework, we evaluated our framework on the real-time perception tasks of StreamingBench, and the backward tracing and real-time understanding tasks of OVO-Bench. The results (\cref{tab:online_video_benchmarks}) show that our framework achieves results comparable to state-of-the-art (SOTA) streaming MLLMs which explicitly optimized context management and real-time understanding. Overall, these results show that our model maintains the online visual understanding abilities of the responding model.
\input{tables/realtime}
\subsection{Computational efficiency}
\begin{figure}[h]
    \centering
    \includegraphics[width=\linewidth]{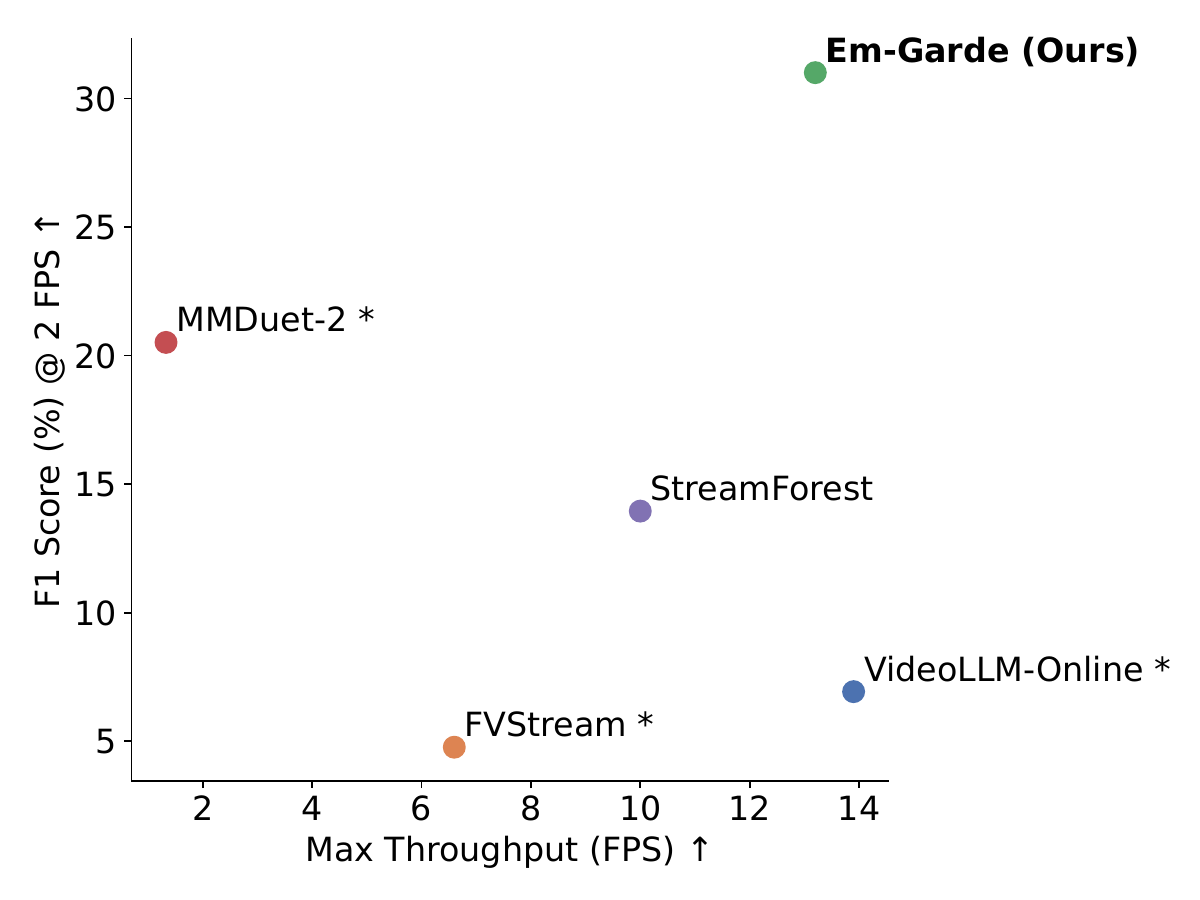}
    \caption{Throughput-Performance comparison on the Proactive Streaming Understanding task. The performance is measured by the average F1 Score on OVO-Bench. Models with * means that the throughput degrade as the context grows without KV Cache truncation. We show the throughput without degrading.}
    \label{fig:efficiency}
\end{figure}
\cref{fig:efficiency} shows the efficiency and performance of state-of-the-art Streaming VideoLLMs on the proactive streaming understanding task. Our model achieves a competitive efficiency of around 13 fps, while outperforming all former models. Furthermore, unlike VideoLLM-Online and MMDuet-2, our model does not suffer from degrading efficiency as the video context grows.
\subsection{Ablation Studies}

\subsubsection{Ablations on IGPP and LPMM}
\input{tables/ablation_arch}
To isolate the contribution of our designed modules, we replace IGPP with direct prompt rewriting using Qwen2.5VL-7B and replace LPMM with a direct yes/no decision module using Qwen2VL-2B (VLM backbone of Ops-MM-V1-2B) We additionally evaluate a naive sliding-window baseline that directly predicts whether to answer from the current query and 2-second video segment. Results are shown in \cref{tab:structure_ablation}. Removing either IGPP or LPMM causes substantial degradation, while removing both components performs even worse. The sliding-window baseline also underperforms despite using the same underlying models. These results validate that the gains of Em-Garde arise from the proposed propose-match decomposition rather than pretrained model capacity alone.

\subsubsection{Triggering Threshold Choices}
We analyze the robustness of under different threshold settings during RL training and inference.Across threshold combinations between 0.03 and 0.04, Em-Garde consistently maintains strong performance on OVO-Bench, with all settings substantially outperforming prior methods. This robustness arises from two factors: (1) triggering is based on changes in similarity rather than absolute similarity values, which is stable across different videos, and (2) RL training adapts proposals to the operating threshold. Therefore, the framework is not overly sensitive to a single manually selected threshold.
\input{tables/threshold_robustness}
\subsubsection{RL coefficient $\lambda$}
\input{tables/reward_ablation}
We additionally study the false-positive penalty coefficient $\lambda$ used during RL training.As shown in \cref{tab:fp_ablation_avg},consistent trends are observed: performance improves when moderate false-positive control is introduced and peaks around $\lambda$ = 1, while both $\lambda$ = 0 (no FP suppression) and overly large $\lambda$ values lead to degraded performance. This suggests that moderate control of false triggers is important for learning perception-friendly proposals. Nevertheless, the performance is relatively stable across $\lambda\in[0.5,1]$, indicating that the performance is not too sensitive to a single $\lambda$.

\section{Limitations}
Despite its effectiveness, Em-Garde has several limitations that point to promising directions for future work.

First, our triggering mechanism relies on thresholding surges in temporal signals, which is sensitive to uncertainty in streaming video. Sudden scene changes or visually correlated yet semantically irrelevant transitions can lead to unstable triggering behavior. We explored suppressing such errors using negative proposals, but current embedding models struggle to reliably distinguish subtly different textual cues. Developing better triggering criteria is an important direction for future research, and we also expect future multimodal embedding models—an active research area—to provide stronger discriminative capabilities.

Second, our approach does not explicitly address online video understanding challenges such as long-horizon reasoning. Following prior work~\cite{dispider}, we treat triggering decision as a problem decoupled from response generation. Em-Garde focuses on \textit{when} to respond, and its modular design allows state-of-the-art responding model to be plugged in downstream. Joint optimization across decision and generation stages remains an open direction for future work.

\section{Conclusion}
We introduced \textbf{Em-Garde}, which separates query parsing from streaming perception to overcome the efficiency–accuracy dilemma in proactive video understanding. By shifting heavy reasoning outside the streaming loop, we enabled a lightweight model to handle triggering decisions accurately and efficiently.

\section*{Acknowledgements}
This work was supported in part by the Tsinghua University (AIR)–AsiaInfo Technologies (China), Inc. Joint Research Center for 6G Network and Intelligent Computing and the Xiongan AI Institute. This work was also supported in part by the National Natural Science Foundation of China (NSFC) under Grants No. 62432004 and No. 62302207, and by the Fundamental Research Funds for the Central Universities under Grants No. 2026300278 and No. ZZKT2026A31.

The authors express their sincere gratitude to all organizations and individuals whose support and contributions made this research possible.
\nocite{langley00}

\bibliography{example_paper}
\bibliographystyle{icml2026}

\newpage
\appendix
\onecolumn
\section{A Demo of Proactive Streaming Understanding}
\begin{figure*}
    \centering
    \includegraphics[width=\textwidth]{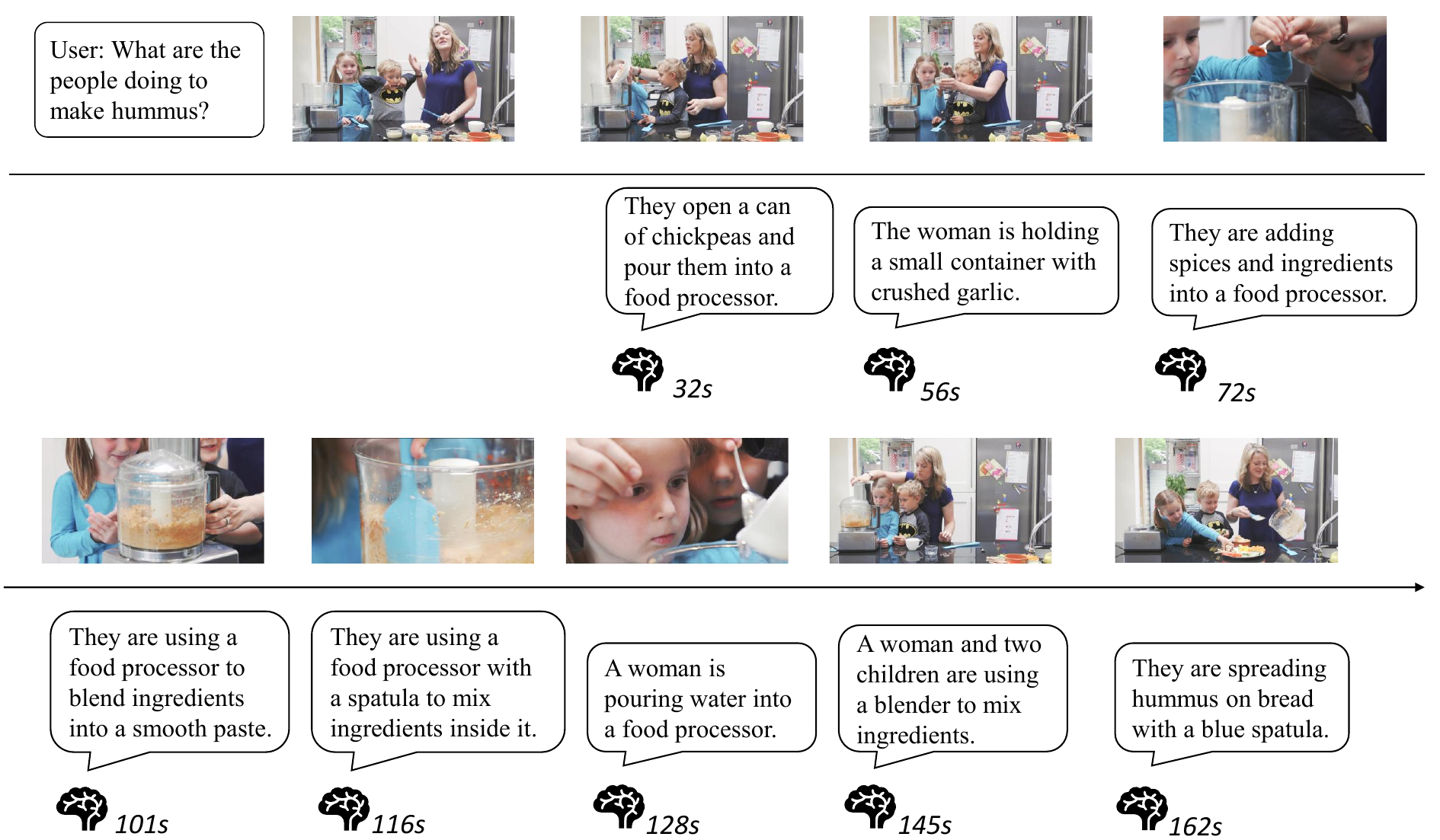}
    \caption{\textbf{A demo of proactive streaming understanding.} In this proactive response task, our model is asked to describe the steps of making hummus in a 3-minute tutorial video. The model correctly identified most of the key steps and gave timely responses, while keeping silent when no new events occur or the content is irrelevant.}
    \label{fig:demo}
\end{figure*}
Fig. \ref{fig:demo} shows a demo of our model's proactive responding performance on a 3-minute tutorial video. The tutorial shows a woman and two kids making hummus in a kitchen, with more than 10 steps and some irrelevant scenes. The model is instructed to detect the steps of making hummus and respond when each new step appears. Our model detects 35 responding timesteps, with 27 accepted by downstream MLLM doublecheck. Among the 27 responses, 12 are timely and accurate description of key events, 10 are duplications of already answered steps, while 5 are wrong or irrelevant. Most key steps are correctly detected by our model within 2 seconds. Overall, our model shows strong ability to proactive respond at key steps while mostly remain silent when no new event appears or the current content is irrelevant. This demonstrates the strong potential for in-the-wild deployment of our method in proactive streaming understanding.
\section{More Results and Analysis}
\subsection{Evaluation Details and Full Results on OVO-Bench Proactive Tasks}
\label{sec:ovo_proactive}
The proactive response tasks (originally called Forward Active Responding) in OVO-Bench include three tasks: Clues Reveal
Responding (CRR) , Sequential Steps
Recognition (SSR) and Repetition Event
Count (REC). The task details are as follows:
\begin{itemize}
    \item \textbf{[REC] Repetition Event
Count.}\quad  Respond when a
repetitive event occurs again, including both high frequency
repetitive actions over short durations and semantically
long-term repetitive occurrences of certain
events.
    \item \textbf{[SSR] Sequential Steps Recognition.}\quad Respond when a
certain procedure or sequence of actions has transitioned
to another stage.
    \item \textbf{[CRR] Clues Reveal Responding.}\quad Respond when sufficient information or clues for a question are provided.
\end{itemize}
Upon the release of OVO-Bench, there were few proactive streaming VideoLLMs that satisfy the original evaluation purposes, so it instead proposed a method to prompt VideoLLMs densely along the time axis to decide whether the current time is appropriate for answering the user’s query. Recent Streaming VideoLLMs mostly follow this scheme and reports the accuracy of the model's answer to the prompt.

\begin{figure}
    \centering
    \includegraphics[width=0.7\linewidth]{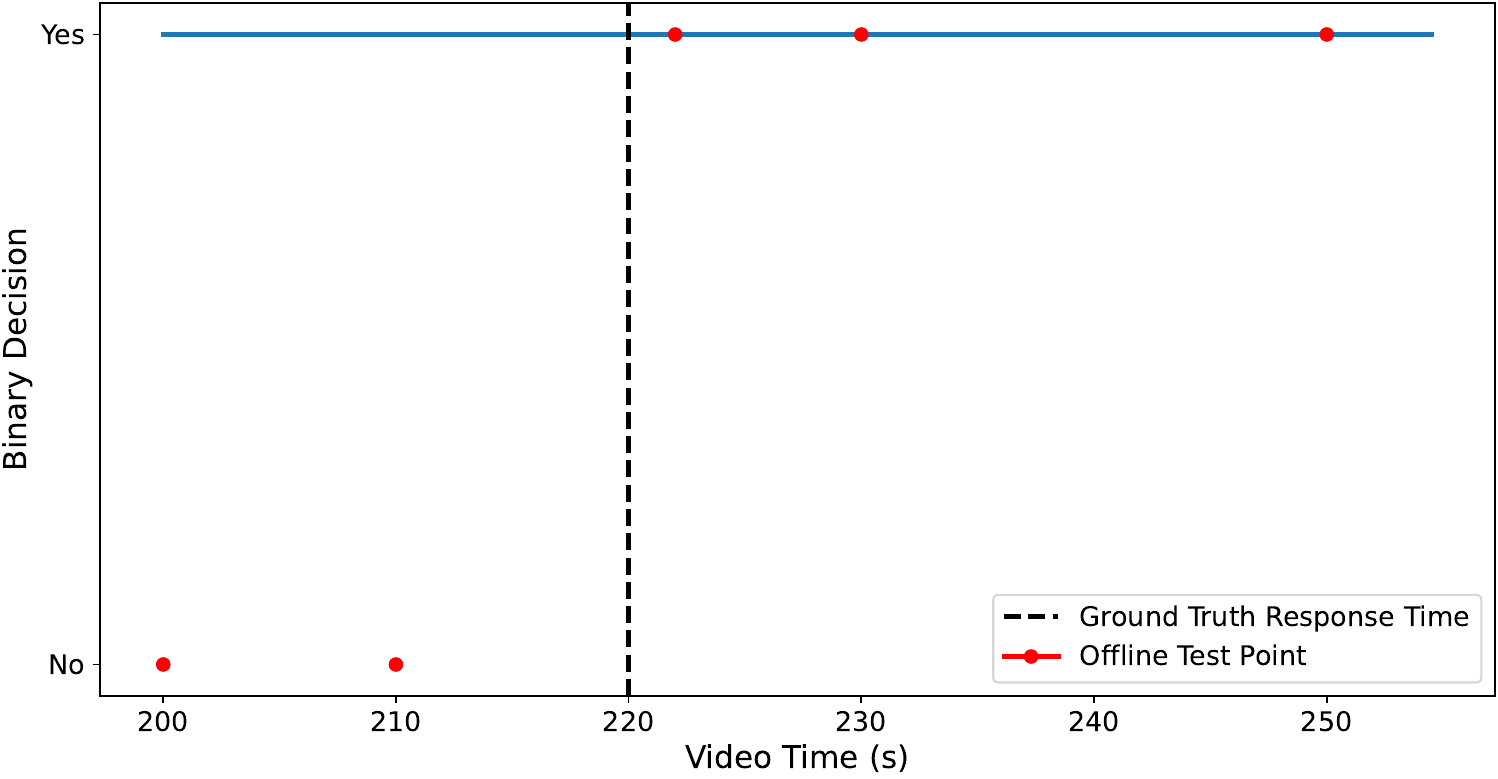}
    \caption{\textbf{Example of a degenerate policy that achieves high offline accuracy on OVO-Bench}. We prompt StreamForest~\cite{zeng2025streamforest} at 2 fps to predict whether the information needed to answer the query is ready, and it always answers "yes". OVO-Bench reports the answer accuracy at the five red timesteps, among which StreamForest answers three correctly (60\% accuracy). However, we cannot make a correct triggering decision from these answers.}
    \label{fig:degenerate}
\end{figure}

However, we notice that this accuracy is biased for evaluating the correctness of triggering decisions. There exist degenerate policies which achieve high  (> 50\%) accuracy but never makes a correct triggering decision, especially for CRR and SSR. For example, in CRR, the model is prompted whether the current information is enough for answering the question. A trigger should be issued when the model's answer changes from "no" to "yes". However, a model which always answer "yes" can achieve 60\% accuracy in the benchmark, while never making a triggering decision (\cref{fig:degenerate}). This clearly illustrates a mismatch between what the metric measures and what it actually aims to evaluate.

As more powerful proactive streaming VideoLLMs emerge, we argue that it is time to return to the original evaluation purpose and explicitly evaluate model's abilities to make triggering decisions. Proactive Streaming VideoLLMs should be evaluated from two dimensions: 1. Whether they can answer at appropriate moments 2. Whether their responses are correct. In this work, we focus on the former dimension.
We introduce a online recall and precision metric: First, we obtain the response time set $T_r$ of the Proactive Streaming VideoLLM. Then we compare it with the ground truth response time set $T_{gt}$. The set of correct response $T_c$ consists of response times in $T_r$ which is within a 2-second window of one response in $T_{gt}$. Finally we compute the recall and precision:
\begin{equation}
    Rec=\frac{|T_c|}{|T_{gt}|},\ Pre=\frac{|T_c|}{|T_r|}
\end{equation}

The recall evaluates whether the model responds at appropriate moments, while the precision evaluates whether it correctly remians silent when no response is required. Together, the metric evaluates the triggering decision ability comprehensively.

For Proactive Streaming VideoLLMs like VideoLLM-Online, MM-Duet2 and Em-Garde (our model), we can directly obtain $|T_r|$. However, many other Streaming VideoLLMs lack their own triggering decision mechanisms. For these models, we followed the prompting method of OVO-Bench, but instead of reporting the offline accuracy of the answer to the query, we obtain $|T_r|$ from the answers (a response is triggered when the answer of the model changes from "no" to "yes" in CRR and SSR, or increase by one in REC) and evaluate the recall and precision of the response times. As shown in \cref{tab:ovo-online}, although these models (FVStream and StreamForest) achieve higher than 50\% answer accuracy, the recall and precision for the triggering decisions degrade significantly. This shows that the accuracy metric is indeed biased, and that the prompting method is inadequate to solve the triggering decision problem.

The detailed recall-precision results are shown in \cref{tab:ovo-online}. We achieve around 50\% recall on all three tasks, significantly higher than previous methods. This means that we can correctly detect and trigger responses at half of the ground truth response times. The precision is relatively lower (but still better than previous methods), meaning that our model still suffers from spurious triggers. Overall, the result is significantly better than previous models, but there is still space for improvement towards a stable, practical solution.

\subsection{Full Results on Online VideoQA Tasks}
\cref{tab:ovo} and \cref{tab:streamingbench_full} show the full evaluation results on Online VideoQA tasks of OVO-Bench and StreamingBench. We achieve comparable or better results on most of the tasks than state-of-the-art (SOTA) Streaming VideoLLMs. This shows that we preserve the online video understanding abilities of the VideoLLM. Together with our triggering decision abilities, it shows that our model has high potential in addressing real-world proactive streaming video understanding tasks.
\input{tables/ovo}
\input{tables/streambench}

\subsection{Inference Efficiency on Long Videos}
\begin{figure}
    \centering
    \includegraphics[width=0.7\linewidth]{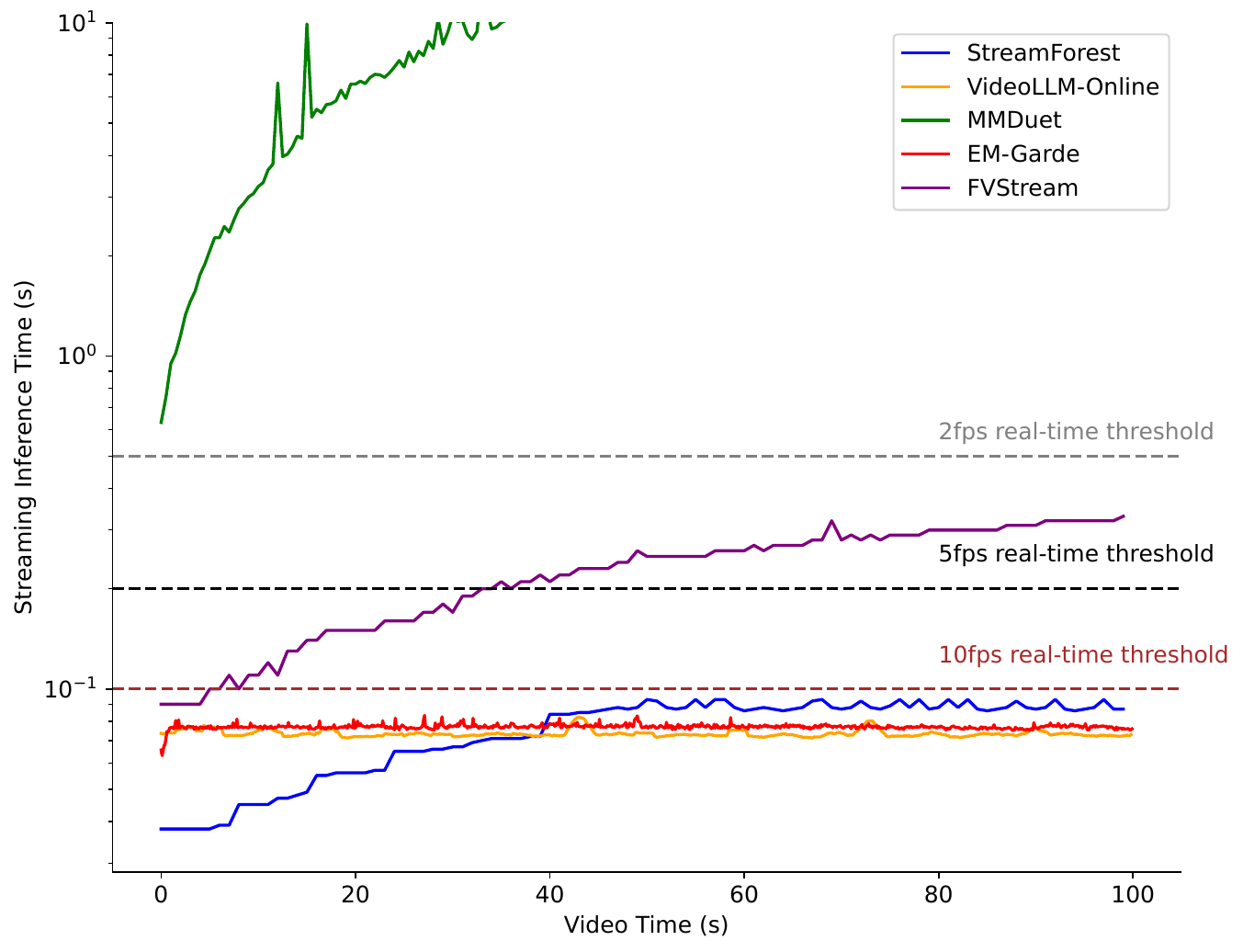}
    \caption{\textbf{Streaming inference latency over video duration in logarithmic scale}}
    \label{fig:latency}
\end{figure}
We compare the streaming inference time of different methods as the video length increases. As shown in \cref{fig:latency}, StreamForest~\cite{zeng2025streamforest} and Em-Garde (ours) maintain nearly constant latency throughout the stream, demonstrating stable real-time performance. VideoLLM-Online~\cite{vl-online} theoretically exhibits linearly increasing latency with video length due to its growing context, although the growth rate is sufficiently small that the effect is barely observable within the evaluated time span. MM-Duet2~\cite{wang2025mmduet2}, while achieving accuracy closest to our method, incurs substantially higher latency that quickly exceeds real-time thresholds, making it impractical for real-time deployment.
\subsection{Analysis of IGPP proposals before and after RL}
\begin{figure}[t]
    \centering
    \includegraphics[width=0.9\linewidth]{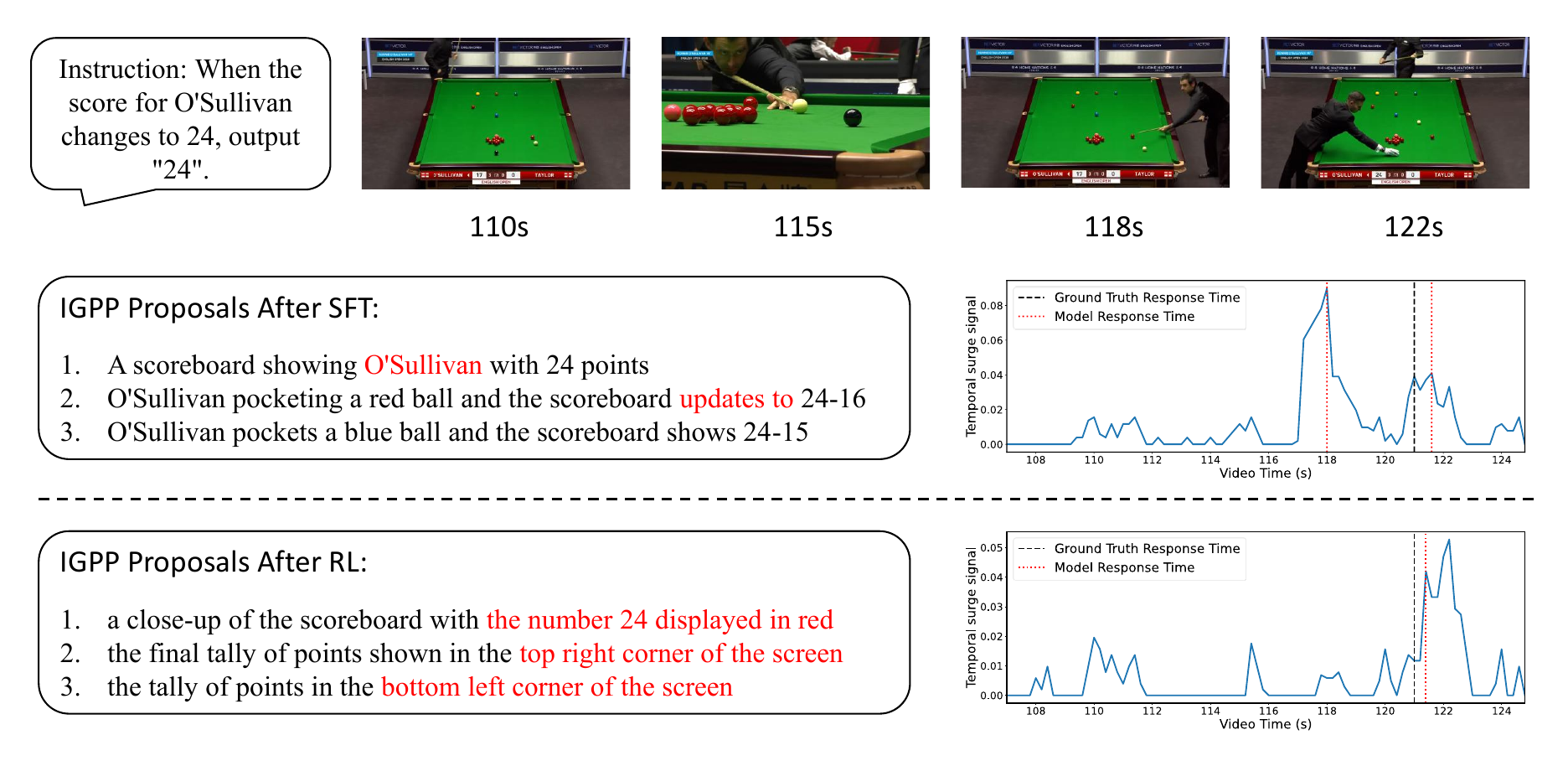}
    \caption{\textbf{Comparison between proposal content (bottom left) and triggering behavior (bottom right) after SFT and RL.} The proposals given by the IGPP before and after RL are shown on the bottom left. The increase of LPMM similarity score and the response times are shown on the bottom right. Due to the different IGPP behaviors, LPMM successfully found a match at the target timestamp for the proposals after RL, while matches at a wrong timestamp (118s) for the proposals before RL.
    }
    \label{fig:ablation1}
\end{figure}
\begin{figure}[h]
    \centering
    \includegraphics[width=\linewidth]{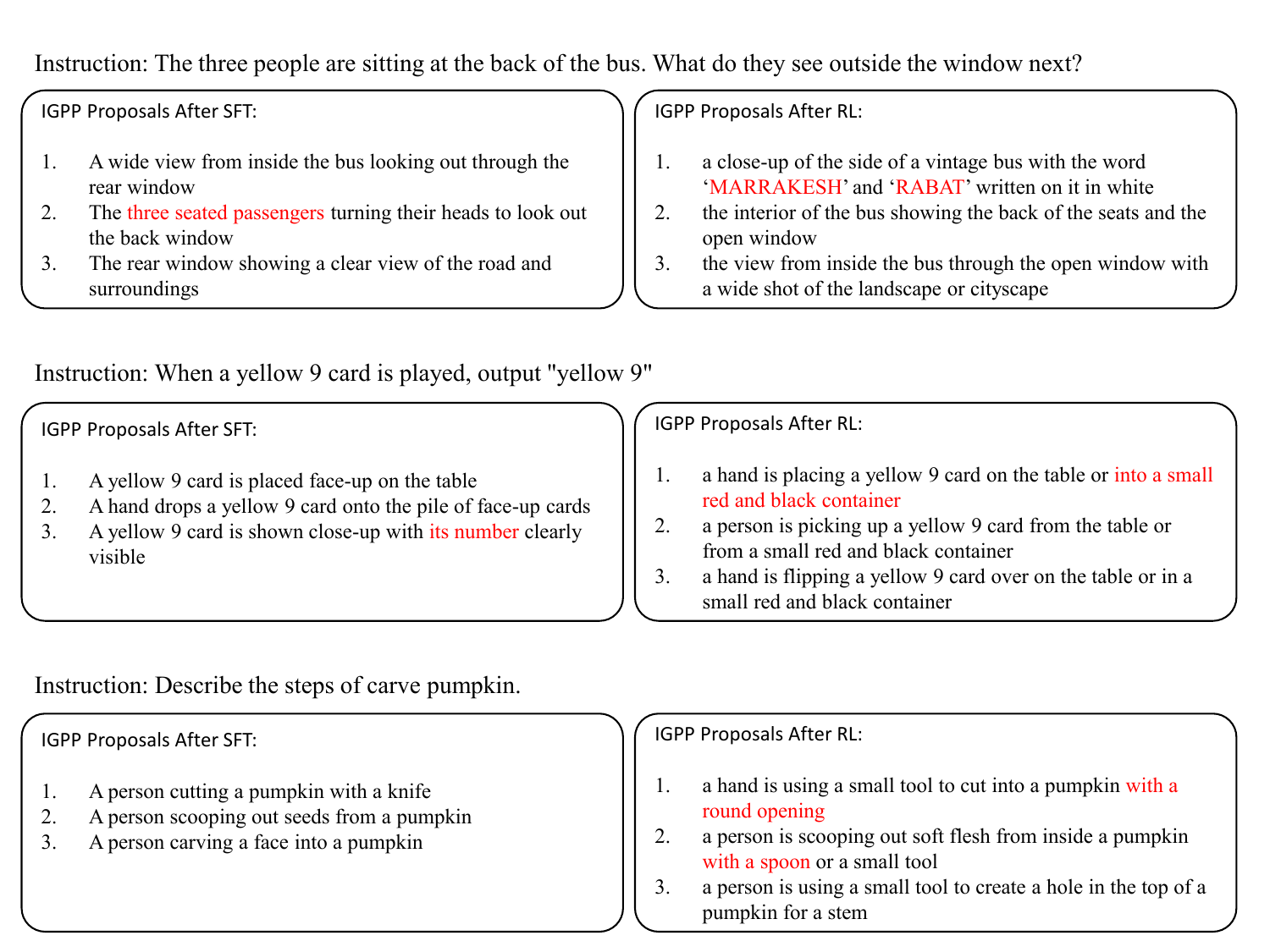}
    \caption{More examples of comparison of IGPP proposals before and after RL}
    \label{fig:proposals}
\end{figure}
We analyze how proposals evolve from supervised fine-tuning (SFT) to reinforcement learning (RL) and how they affects triggering behavior. As an example shown in \cref{fig:ablation1},  after SFT, proposals are already aligned to the target events, but they might focus on concepts which not only appear when the response should be given (poor temporal alignment) or too abstract for the perception module (poor perceptual groundability). For example, in the query “When the score for O'Sullivan changes to 24, output ‘24’”, all proposals given by the model before RL include “O’Sullivan”, which appears often in the video and distracts the LPMM. While plausible at a semantic level, these proposals can confuse the LPMM, making the model sensitive to irrelevant scene changes.

After RL, proposals consistently shift toward temporal-aligned, perceptual-grounded visual cues. They focus on specific objects and their interactions which appears exactly when the response should be given (e.g. the number 24 displayed in red) and improve redundancy by listing different viewpoints and layouts. This shows that by learning which visual cues suit the LPMM better, the model learns to generate proposals that can effectively guide triggering decisions. Additional qualitative examples and analyses are provided in the supplementary material.
\cref{fig:proposals} show more examples of IGPP proposals before and after RL. The comparison show that after RL, IGPP consistently produce more temporal-aligned and perceptual-grounded proposals. The redundancy of the proposals is improved by enumerating possible detailed actions, viewpoints and objects. The IGPP also utilize the sampled video frames better to include objects and background information which is not in the query, but can help make trigger decisions (e.g. "MARRAKESH", "red and black container").
\section{Details of the training Dataset}
\label{sec:dataset}
Our training dataset, Parse2Props-1K, contains 668 instructions drawn from 92 videos. The videos are randomly selected from COIN~\cite{coin}, Ego4D~\cite{grauman2022ego4d} and BEHAVIOR~\cite{li2024behavior1k} dataset.
\paragraph{COIN} COIN is a dataset of youtube videos with step-level annotations. It consists of 11,827 videos related to 180 different tasks, including tutorials, daily activities, and advertisements. It features a wide range of events and activities, and in-the-wild daily life scenarios.

\paragraph{Ego4D} Ego4D is the world's largest egocentric (first person) video ML dataset and benchmark suite, featuring long action sequences, dense narration, and wide range of scenes. It covers various scenarios (household, outdoor, workplace, leisure, etc.) of daily life activity captured in-the-wild by 926 camera wearers from 74 worldwide locations and 9 different countries.
\paragraph{BEHAVIOR} BEHAVIOR-1K is a comprehensive simulation benchmark for testing embodied AI agents on 1,000 everyday household activities. It contains RGB videos of robot action trajectories collected from the head and waist camera. We use the RGB videos from the head camera as our raw videos. 

For dataset annotations, we first designed 40 instructions that the model needs to respond to proactively (i.e., the clue appears after the query time). The instructions span 7 tasks: Object Recognition, Attribute Recognition, Action Recognition, Spatial Reasoning, Temporal Reasoning, Step Recognition and Clue Recognition. The task definitions are:
\paragraph{Object Recognition} The model needs to give a response when it finds a certain object.
\paragraph{Action Recognition} The model needs to give a response when a certain action is performed.
\paragraph{Attribute Recognition} The model needs to give a response when an object displays a certain visual attribute.
\paragraph{Spatial Reasoning} The model needs to give a response when a certain spatial relationship between objects is observed.
\paragraph{Temporal Reasoning} The model is asked what happens after a certain activity, and needs to respond when the following activity occurs.
\paragraph{Step Recognition} The model needs to give a response when each new step related to the given task is performed.
\paragraph{Clue Recognition} The model needs to respond when a clue that warrants an answer to the given open question appears.

For each human-written proposal, we also provide the query time, target response time, and a set of proposals. The proposals are written with a set of basic proposing methods and rules, including anticipating the target scene using common knowledge, changing high-level ideas to visual details, and adding related visual context. 

\begin{figure}[tb]
    \centering
    \includegraphics[width=\linewidth]{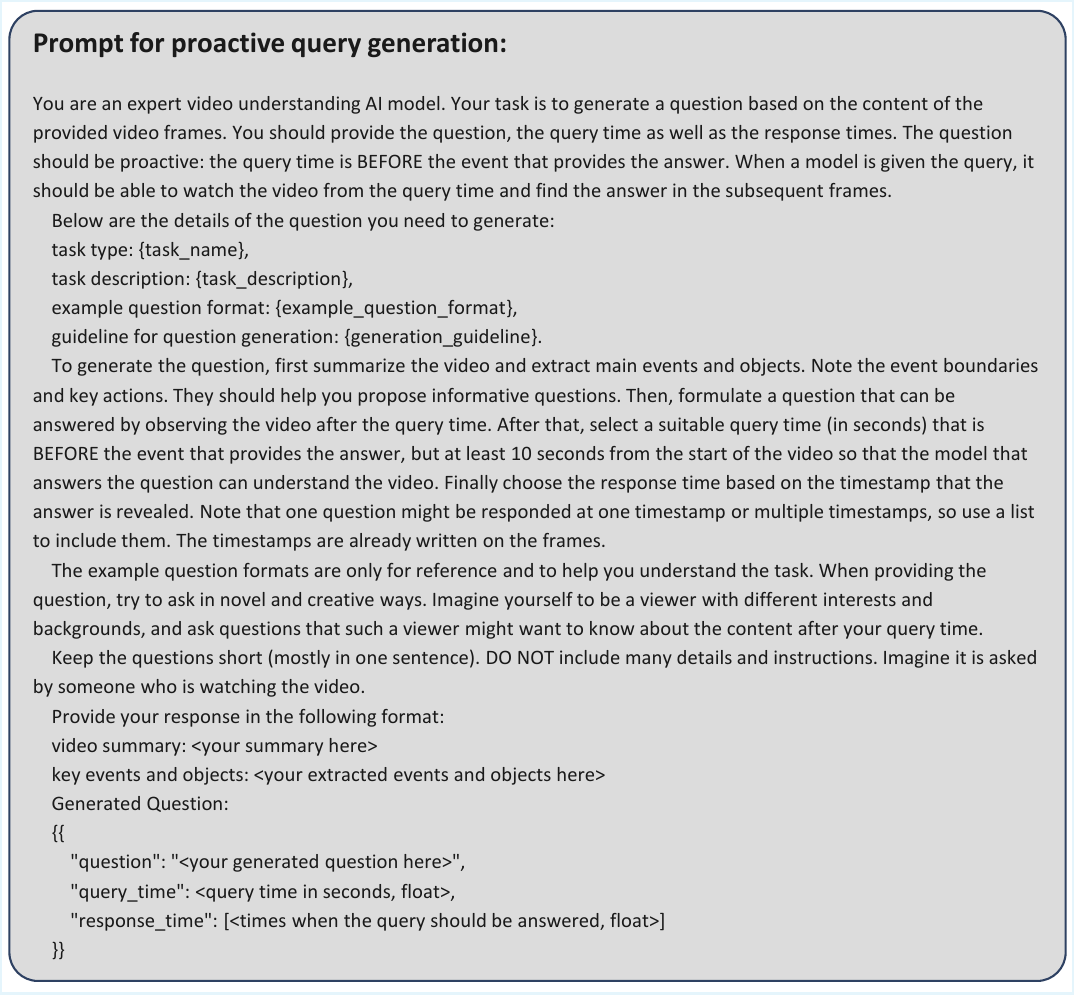}
    \caption{Prompt for proactive query generation}
    \label{fig:prompt1}
\end{figure}

\begin{figure}[tb]
    \centering
    \includegraphics[width=\linewidth]{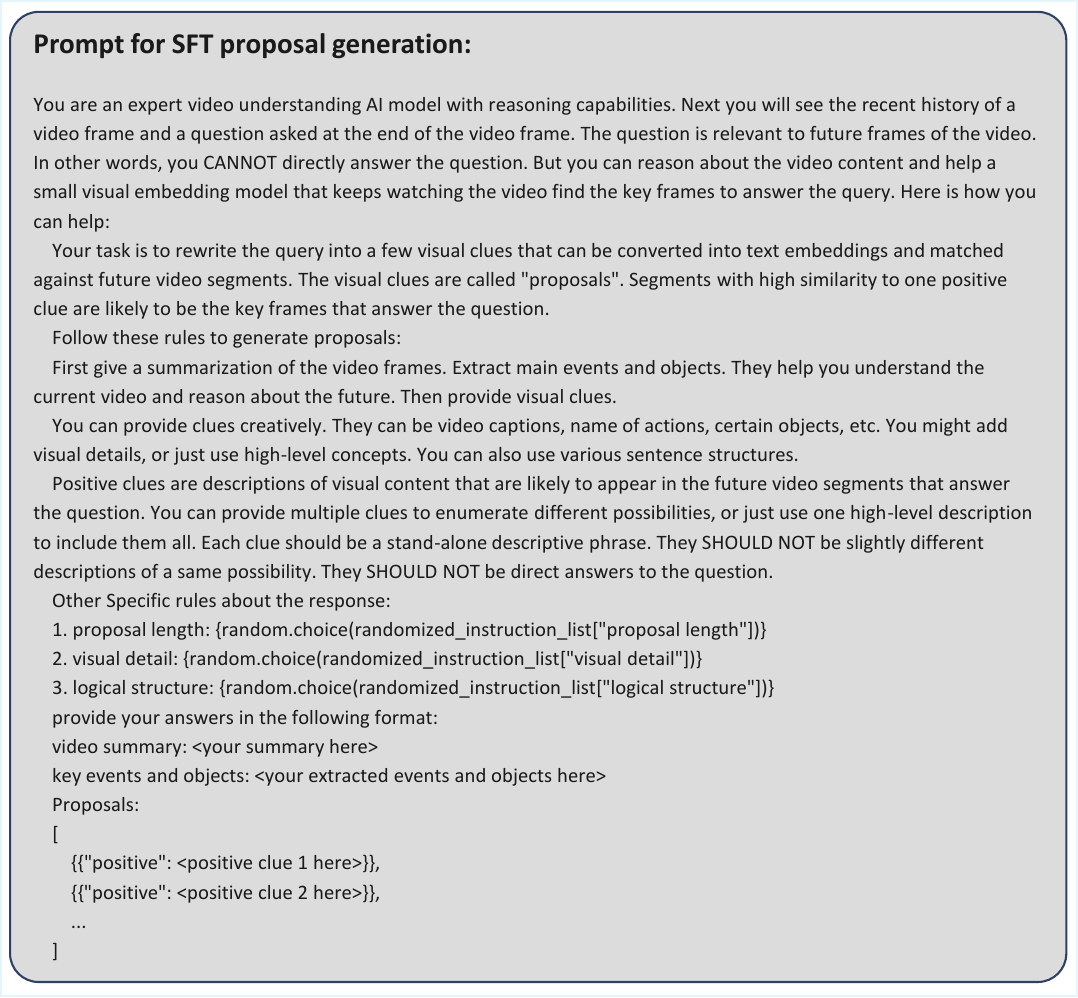}
    \caption{Prompt for SFT Proposal generation}
    \label{fig:prompt2}
\end{figure}

\begin{figure}[tb]
    \centering
    \includegraphics[width=\linewidth]{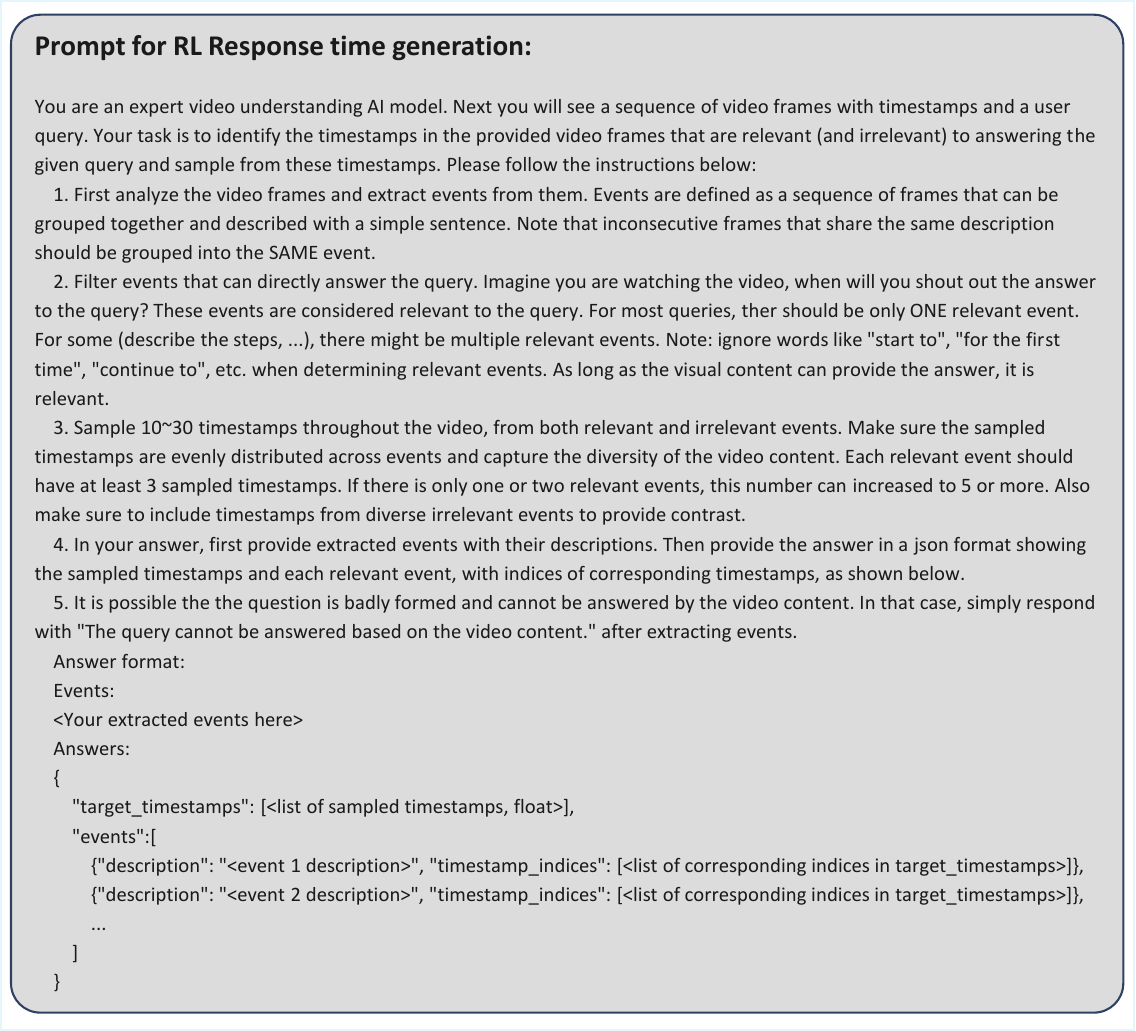}
    \caption{Prompt for RL Response time generation}
    \label{fig:prompt3}
\end{figure}

After annotating the 40 instructions-proposal pairs by hand, we prompt GPT-5 to generate more instructions and proposals for each task. We first clip the sampled videos from the datasets to segments of 30 seconds, and ask GPT-5 to ask a question at a certain time that needs future content to answer.These questions are used as instructions for proactive streaming understanding. After the instructions are generated, we ask GPT-5 to generate proposals following the hand-written examples. To increase the variety of the proposals, we randomize the prompt for proposal generation along three axes: the level of visual details, the complexity of semantic logic, and the length of the proposals. We also ask GPT-5 to try to answer the questions, and point out the time steps of the frames that are related to the answer. The prompts for instruction, proposal, and response time generation are shown in \cref{fig:prompt1}, \cref{fig:prompt2} and \cref{fig:prompt3}.

Finally we filter the GPT-annotated examples by hand. We remove the questions that (1) cannot be answered given the video segment (2) do not need the video to answer, (3) do not correspond to a clear response time or (4) focus on insignificant details. We kept 668 examples after filtering.

For SFT, we selected 334 well-written proposals in the 668 examples to form instruction-proposal pairs. For RL, we selected 400 well temporal-grounded responses as the ground truth responding time.



\end{document}

%% file: tables/ovo_online.tex
\begin{table*}[t]
\centering
\small
\renewcommand{\arraystretch}{1.2}
\setlength{\tabcolsep}{4pt}  
\begin{tabular}{lcc ccc ccc ccc c}
\toprule
\multirow{2}{*}{Model}
& \multirow{2}{*}{Size}

& \multirow{2}{*}{FPS}
& \multicolumn{3}{c}{CRR} 
& \multicolumn{3}{c}{SSR} 
& \multicolumn{3}{c}{REC} 
& \multirow{2}{*}{Avg. F1} \\
\cmidrule(lr){4-6} \cmidrule(lr){7-9} \cmidrule(lr){10-12}
& & & R & P & F1 & R & P & F1 & R & P & F1 &  \\
\midrule
VideoLLM-Online~\cite{vl-online}& 8B &2 & 2.08 & 2.08 & 2.08 & 11.07 & 6.88 & 8.49 & 7.43 & 16.46 & 10.24 & 6.93 \\
FVStream~\cite{fvstream}& 7B & 2 & 6.25 & 5.21 & 5.68 & 2.46 & 9.52 & 3.91 & 3.46 & 7.49 & 4.73 & 4.77 \\
StreamForest~\cite{zeng2025streamforest}& 7B & 2 & 2.08 & 2.08 & 2.08 & 8.02 & 13.69 & 10.11 & 30.49 & 28.90 & 29.67 & 13.95 \\
MMDuet-2~\cite{wang2025mmduet2}& 3B & 2 & 47.92 & 8.47 & 14.40 & -- & -- & -- & 19.33 & \textbf{42.72} & 26.62 & 20.51 \\
\midrule
Em-Garde (Ours) & 2B & 2 & \textbf{47.92}
 & \textbf{18.22} & \textbf{26.40} & \textbf{54.88} & \textbf{14.87} & \textbf{23.40} & \textbf{47.33} & 39.66 & \textbf{43.16} & \textbf{30.99} \\
\bottomrule
\end{tabular}
\hspace{2pt}
\caption{Proactive Response Timing comparison on OVO-Bench Future Active Responding tasks. Our model significantly outperforms existing proactive streaming models and prompt-based proactive response methods. Dispider~\cite{dispider} and StreamAgent~\cite{yang2025streamagent} are not evaluated due to lack of open-source proactive decision code.}
\label{tab:ovo-online}
\end{table*}

%% file: tables/streambench_online.tex
\begin{table}[t]
\footnotesize
\setlength{\tabcolsep}{4pt}
\centering
\renewcommand{\arraystretch}{1.2}
\begin{tabular}{l c c c}
\hline
Model & Size & FPS & Accuracy \\
\hline
FVStream~\cite{fvstream} & 7B & - & 2.0\\
VideoLLM-Online~\cite{vl-online} & 8B & 2 & 4.0 \\
Dispider~\cite{dispider} & 7B & 1 & 25.3 \\
StreamAgent~\cite{yang2025streamagent} & 3B & 1 & 28.9 \\
MMDuet-2~\cite{wang2025mmduet2} & 3B & 1 & 34.6 \\
\hline
Em-Garde (Ours) & 2B & 1 & 37.6\\
Em-Garde (Ours) & 2B & 2 & \textbf{38.0}\\
\hline
\end{tabular}
\caption{Comparison of models on the accuracy for StreamingBench PO Task. Our model outperforms existing proactive streaming models by more than 3\% accuracy.}
\label{tab:po}
\end{table}

%% file: tables/proactivevideoqa.tex
\begin{table}[t]
\centering
\small
\renewcommand{\arraystretch}{1.2}
\begin{tabular}{lccc}
\hline
Model & WEB & EGO & VAD \\
\hline
VideoLLM-Online & 25.9 / -- & 25.0 / -- &25.0 / -- \\
MMDuet &38.9 / 81.3 & 46.0 / 99.4& 27.4 / 99.2\\
MMDuet-2 & \textbf{53.3} / \textbf{4.2} & 33.6 / \textbf{8.1} & \textbf{28.9} / 15.2\\
\hline
Em-Garde(Ours) & 44.3 /4.5 & \textbf{52.3} / 17.4& 27.4 / \textbf{1.4}\\
\hline
\end{tabular}
\caption{Comparison of models on the \textbf{PAUC ($\uparrow$) / reply duplicate proportion ($\downarrow$)} of the ProactiveVideoQA benchmark. We achieve competitive results against MMDuet-2, which \textit{directly} optimizes the PAUC metric and is trained on related datasets.}
\label{tab:proactiveqa}
\end{table}

%% file: tables/realtime.tex
\begin{table}[h]

\centering
\scriptsize
\setlength{\tabcolsep}{1.5pt}
\renewcommand{\arraystretch}{1.05}
\begin{tabular}{l c c c c}
\toprule
Model & Size (B) 
& StreamingBench 
& \multicolumn{2}{c}{OVO-Bench} \\
\cmidrule(lr){3-3} \cmidrule(lr){4-5} 
 &  
 & Real-time VU $\uparrow$ 
 & Real-time VP $\uparrow$ 
 & Backward Tracing $\uparrow$ \\
\midrule
VideoLLM-Online & 8B & 36.0 & 20.8 & 17.7 \\
FVStream & 7B & 23.2 & 29.9 & 25.4 \\
Dispider & 7B & 67.6 & 54.6 & 36.1 \\
StreamAgent & 7B & 74.3 & \underline{61.3} & 46.2 \\
StreamForest & 7B & \textbf{77.3} & 61.2 & \underline{52.0} \\
\midrule
Em-Garde (Ours) & 7B & \underline{76.7} & \textbf{63.0} & \textbf{52.2} \\
\bottomrule
\end{tabular}
\caption{Performance comparison on online video understanding benchmarks. Higher is better for all metrics.}
\label{tab:online_video_benchmarks}
\end{table}

%% file: tables/ablation_arch.tex
\begin{table}[t]
\centering
\scriptsize
\renewcommand{\arraystretch}{1}
\setlength{\tabcolsep}{6pt}
\begin{tabular}{lcc}
\toprule
Method & OVO-Bench (FAR) & StreamingBench (PO) \\
\midrule
Em-Garde          & \textbf{31.0} & \textbf{38.0} \\
w/o. IGPP   & 24.0 & 28.8 \\
w/o. LPMM & 14.6 & 23.2 \\
w/o. IGPP \& LPMM   & 18.6 & 17.2 \\
sliding window & 21.0 & 26.8 \\
\bottomrule
\end{tabular}
\vspace{2pt}
\caption{Ablation on IGPP \& LPMM. Without our propose-match design, the performance significantly degrades.}
\label{tab:structure_ablation}
\end{table}

%% file: tables/threshold_robustness.tex
\begin{table}[t]
\centering
\small
\label{tab:threshold_robustness}
\begin{tabular}{lccc}
\toprule
Train $\theta$ $\backslash$ Test $\theta$
& 0.03 & 0.035 & 0.04 \\
\midrule
0.03  & 26.59 & 26.89 & 26.67 \\
0.035 & \textbf{28.74} & \textbf{30.12} & 29.44 \\
0.04  & 27.25 & 28.24 & \textbf{30.99} \\
\bottomrule
\end{tabular}
\caption{OVO-Bench average F1 score under different training and testing thresholds.}
\end{table}

%% file: tables/reward_ablation.tex
\begin{table}[h]
\centering
\small
\setlength{\tabcolsep}{6pt}
\begin{tabular}{lcc}
\toprule
Method & OVO-Bench (FAR) & StreamingBench (PO) \\
\midrule
SFT                & 23.4 & 34.8 \\
RL ($\lambda=0$)   & 27.7 & 30.8 \\
RL ($\lambda=0.5$) & 28.2 & 33.2 \\
RL ($\lambda=0.75$) & 28.6 & 36.4 \\
RL ($\lambda=1$)   & \textbf{31.0} & \textbf{38.0} \\
RL ($\lambda=1.5$) & 19.3 & 33.6 \\
\bottomrule
\end{tabular}
\vspace{2pt}
\caption{
Ablation of the false-positive penalty coefficient $\lambda$. The OVO-Bench score is the overall F1 score for Forward Active Responding tasks. 
}
\label{tab:fp_ablation_avg}
\end{table}

%% file: tables/ovo.tex
\begin{table*}[t]
\centering
\small
\setlength{\tabcolsep}{3.3pt}
\renewcommand{\arraystretch}{1.1}

\begin{tabular}{l c | cccccc c | ccc c}
\toprule
\multirow{2}{*}{Model} & \multirow{2}{*}{\# Frames} 
& \multicolumn{7}{c|}{Real-Time Visual Perception} 
& \multicolumn{4}{c}{Backward Tracing}   \\
\cmidrule(lr){3-9} \cmidrule(lr){10-13}
& 
& OCR & ACR & ATR & STU & FPD & OJR & Avg.
& EPM & ASI & HLD & Avg. \\
\midrule

\multicolumn{13}{c}{\textbf{Human}} \\
\midrule
Human Agents & -- &
94.0 & 92.6 & 94.8 & 92.7 & 91.1 & 94.0 & 93.2 & 92.6 & 93.0 & 91.4 & 92.3  \\
\midrule

\multicolumn{13}{c}{\textbf{Proprietary Multimodal Models (Offline)}} \\
\midrule
Gemini~1.5~Pro~\cite{team2024gemini}
& 1fps
& 87.3 & 67.0 &  80.2 & 54.5 & 68.3 & 67.4 & 70.8 & 68.6 & 75.7 & 52.7 & 62.3 \\

GPT-4o~\cite{hurst2024gpt}
& 64
& 69.1 & 65.1 & 65.5 & 50.0 & 68.3 & 63.7 & 63.6 & 49.8 & 71.0 & 55.4 & 58.7 \\
\midrule

\multicolumn{13}{c}{\textbf{Open-source Multimodal Models (Offline)}} \\
\midrule
LLaVA-Video-7B~\cite{llavavideo}
& 64
& 69.8 & 59.6 & 66.4 & 50.6 & 72.3 & 61.4 & 63.3
& 51.2 & 64.2 & 9.7 & 41.7 \\

LLaVA-OneVision-7B~\cite{llava-onevision}
& 64
& 67.1 & 58.7 & 69.8 & 49.4 & 71.3 & 60.3 & 62.8
& 52.5 & 58.8 & 23.7 & 45.0 \\

Qwen2-VL-7B~\cite{qwen2}
& 64
& 69.1 & 53.2 & 63.8 & 50.6 & 66.3 & 60.9 & 60.7
& 44.4 & 66.9 & 34.4 & 48.6 \\

InternVL-V2-8B~\cite{internvl2}
& 64
& 68.5 & 58.7 & 69.0 & 44.9 & 67.3 & 56.0 & 60.7
& 43.1 & 61.5 & 27.4 & 44.0 \\

LongVU-7B~\cite{longvu}
& 1fps
& 55.7 & 49.5 & 59.5 & 48.3 & 68.3 & 63.0 & 57.4
& 43.1 & 66.2 & 9.1 & 39.5 \\

\midrule
\multicolumn{13}{c}{\textbf{Open-source Multimodal Models (Online)}} \\
\midrule
Flash-VStream-7B~\cite{fvstream}
& 1fps
& 25.5 & 32.1 & 29.3 & 33.7 & 29.7 & 28.8 & 29.9
& 36.4 & 33.8 & 5.9 & 25.4 \\

VideoLLM-online-8B~\cite{vl-online}
& 2fps
& 8.1 & 23.9 & 12.1 & 14.0 & 45.5 & 21.2 & 20.8
& 22.2 & 18.8 & 12.2 & 17.7 \\

Dispider~\cite{dispider}
& 1fps
& 57.7 & 49.5 & 62.1 & 44.9 & 61.4 & 51.6 & 54.5
& 48.5 & 55.4 & 4.3 & 36.1 \\

StreamAgent-7B~\cite{yang2025streamagent}
& 1fps
& \underline{71.2} & \underline{53.2} & 63.6 & \textbf{53.9} & \underline{67.3} & \underline{58.7} & \underline{61.3}
& \underline{54.8} & 58.1 & 25.8 & 46.2 \\

StreamForest-7B~\cite{zeng2025streamforest}
& 1fps
& 68.5 & \underline{53.2} & \textbf{71.6} & \underline{47.8} & 65.4 & \textbf{60.9} & 61.2
& \textbf{58.9} & \textbf{64.9} & \underline{32.3} & \underline{52.0} \\
\midrule
\textbf{Em-Garde-7B (Ours)}
& 1fps
& \textbf{76.5} & \textbf{61.5} & \underline{65.5} & 47.2 & \textbf{70.3} & 57.1 & \textbf{63.0}
& 48.2 & \underline{63.5} & \textbf{45.2} & \textbf{52.2} \\
\bottomrule
\end{tabular}

\caption{Performance comparison on Real-time understanding and Backward Tracing tasks of OVO-Bench}
\label{tab:ovo}
\end{table*}

%% file: tables/streambench.tex
\begin{table*}[t]
\centering
\small
\setlength{\tabcolsep}{3.2pt}
\begin{tabular}{lccccccccccccc}
\toprule
Method & Size & \# Frames & OP & CR & CS & ATP & EU & TR & PR & SU & ACP & CT & ALL \\
\midrule
Human & -- & -- 
& 89.5 & 92.0 & 93.6 & 91.5 & 95.7 & 92.5 & 88.0 & 88.8 & 89.7 & 91.3 & 91.5 \\
\midrule
\multicolumn{14}{l}{\textit{Proprietary MLLMs}} \\
Gemini 1.5 pro~\cite{team2024gemini} & -- & 1fps 
& 79.0 & 80.5 & 83.5 & 79.7 & 80.0 & 84.7 & 77.8 & 64.2 & 72.0 & 48.7 & 75.7 \\
GPT-4o~\cite{hurst2024gpt} & -- & 64
& 77.1 & 80.5 & 83.9 & 76.5 & 70.2 & 83.8 & 66.7 & 62.2 & 69.1 & 49.2 & 73.3 \\
Claude 3.5 Sonnet~\cite{claude} & -- & 20
& 73.3 & 80.5 & 84.1 & 82.0 & 75.4 & 79.5 & 61.1 & 61.8 & 69.3 & 43.1 & 72.4 \\
\midrule
\multicolumn{14}{l}{\textit{Open-source Offline Video MLLMs}} \\
Video-LLaMA2~\cite{cheng2024videollama2} & 7B & 32
& 55.9 & 55.5 & 57.4 & 58.2 & 52.8 & 43.6 & 39.8 & 42.7 & 45.6 & 35.2 & 49.5 \\
VILA-1.5~\cite{fang2024vila2} & 8B & 14
& 53.7 & 49.2 & 71.0 & 56.9 & 53.4 & 53.9 & 54.6 & 48.8 & 50.1 & 17.6 & 52.3 \\
Video-CCAM~\cite{fei2024videoccam} & 14B & 96
& 56.4 & 57.8 & 65.3 & 62.8 & 64.6 & 51.4 & 42.6 & 48.0 & 49.6 & 31.6 & 54.0 \\
LongVA~\cite{zhang2024longva} & 7B & 128
& 70.0 & 63.3 & 61.2 & 70.9 & 62.7 & 59.5 & 61.1 & 53.7 & 54.7 & 34.7 & 60.0 \\
InternVL2~\cite{internvl2} & 8B & 16
& 68.1 & 60.9 & 69.4 & 77.1 & 67.7 & 62.9 & 59.3 & 53.3 & 55.0 & 56.5 & 63.7 \\
Kangaroo~\cite{kangaroogroup} & 7B & 64
& 71.1 & 84.4 & 70.7 & 73.2 & 67.1 & 61.7 & 56.5 & 55.7 & 62.0 & 38.9 & 64.6 \\
LLaVA-NeXT-Video~\cite{liu2024llavanext} & 32B & 64
& 78.2 & 70.3 & 73.8 & 76.8 & 63.4 & 69.8 & 57.4 & 56.1 & 64.3 & 38.9 & 67.0 \\
MiniCPM-V2.6~\cite{yao2024minicpm} & 8B & 32
& 71.9 & 71.1 & 77.9 & 75.8 & 64.6 & 65.7 & 70.4 & 56.1 & 62.3 & 53.4 & 67.4 \\
LLaVA-OneVision~\cite{llava-onevision} & 7B & 32
& 80.4 & 74.2 & 76.0 & 80.7 & 72.7 & 71.7 & 67.6 & 65.5 & 65.7 & 45.1 & 71.1 \\
Qwen2.5-VL~\cite{qwen2_5} & 7B & 1fps
& 78.3 & 80.5 & 78.9 & 80.5 & 76.7 & 78.5 & 79.6 & 63.4 & 66.2 & 53.2 & 73.7 \\
\midrule
\multicolumn{14}{l}{\textit{Open-source Online Video MLLMs}} \\
Flash-VStream~\cite{fvstream} & 7B & --
& 25.9 & 43.6 & 24.9 & 23.9 & 27.3 & 13.1 & 18.5 & 25.2 & 23.9 & 48.7 & 23.2 \\
VideoLLM-online~\cite{vl-online} & 8B & 2fps
& 39.1 & 40.1 & 34.5 & 31.1 & 46.0 & 32.4 & 31.5 & 34.2 & 42.5 & 27.9 & 36.0 \\
Dispider~\cite{dispider} & 7B & 1fps
& 74.9 & 75.5 & 74.1 & 73.1 & 74.4 & 59.9 & 76.1 & 62.9 & 62.2 & 45.8 & 67.6 \\
StreamAgent~\cite{yang2025streamagent} & 7B & 1fps
& 79.6 & \underline{78.3} & 79.2 & 75.9 & 74.7 & 76.9 & \underline{82.9} & 66.3 & \underline{73.7} & \textbf{55.4} & 74.3 \\
StreamForest~\cite{zeng2025streamforest} & 7B & 1fps
& \textbf{83.1} & \textbf{82.8} & \underline{82.7} & \textbf{84.3} & \underline{77.5} & \underline{78.2} & 76.9 & \underline{69.1} & \textbf{75.6} & \underline{54.4} & \textbf{77.3} \\
Em-Garde & 7B & 2fps
& \underline{80.4} & 74.2 & \textbf{89.9} & \underline{84.0} & \textbf{78.9} & \textbf{82.2} & \textbf{87.0} & \textbf{71.1} & 68.8 & 42.5 & \underline{76.7} \\
\bottomrule
\end{tabular}
\caption{Full evaluation results of real-time understanding tasks on StreamingBench.}
\label{tab:streamingbench_full}
\end{table*}